\def\paperID{5519}
\def\confName{ICCV}
\def\confYear{2025}

\def\paperTitle{Multi-GraspLLM: A Multimodal LLM for Multi-Hand \\Semantic Guided Grasp Generation}

\def\authorBlock{
Haosheng Li$^{1}$\thanks{Equal contribution.}, \quad Weixin Mao$^{2*}$\thanks{Project Lead.}, \quad Weipeng Deng$^3$,  \quad  Chenyu Meng$^1$,\quad  Haoqiang Fan$^4$, \\
Tiancai Wang$^4$,\quad Yoshie Osamu$^2$,\quad Ping Tan$^5$,\quad Hongan Wang$^1$,\quad Xiaoming Deng$^{1}$\thanks{Corresponding author.}\\
$^1$Institute of Software, Chinese Academy of Sciences\\
$^2$Waseda University $^3$University of Hong Kong
 $^4$MEGVII Technology\\
$^5$Hong Kong University of Science and Technology\\
}

\newif\ifreview 
\newif\ifarxiv \newcommand{\arxiv}{\arxivtrue}
\newif\ifcamera 
\newif\ifrebuttal 

\arxiv
\pdfoutput=1
\documentclass[10pt,twocolumn,letterpaper]{article}
\ifreview \usepackage[review]{cvpr} \fi
\ifarxiv \usepackage[pagenumbers]{cvpr} \fi
\ifrebuttal \usepackage[rebuttal]{cvpr} \fi
\ifcamera \usepackage{cvpr} \fi

\usepackage{pifont}
\newcommand{\cmark}{\ding{51}}%

\usepackage{graphicx}	
\usepackage{amsmath}	
\usepackage{amssymb}	
\usepackage{booktabs}
\usepackage{times}
\usepackage{microtype}
\usepackage{epsfig}
\usepackage{caption}
\usepackage{float}
\usepackage{placeins}
\usepackage{color, colortbl}
\usepackage{stfloats}
\usepackage{enumitem}
\usepackage{tabularx}
\usepackage{xstring}
\usepackage{multirow}
\usepackage{xspace}
\usepackage{url}
\usepackage{subcaption}
\usepackage{xcolor}
\usepackage[hang,flushmargin]{footmisc}

\ifcamera \usepackage[accsupp]{axessibility} \fi

% ...

\ifarxiv  \fi

\newcommand{\R}[1]{{%
    \textbf{%
        \ifstrequal{#1}{1}{\textcolor{red}{R#1}}{%
        \ifstrequal{#1}{2}{\textcolor{blue}{R#1}}{%
        \ifstrequal{#1}{3}{\textcolor{magenta}{R#1}}{%
        \ifstrequal{#1}{4}{\textcolor{teal}{R#1}}{%
                           \textcolor{cyan}{R#1}%
        }}}}%
    }%
}}

\usepackage{xr-hyper}

\makeatletter
\newcommand*{\addFileDependency}[1]{
  \typeout{(#1)}
  \@addtofilelist{#1}
  \IfFileExists{#1}{}{\typeout{No file #1.}}
}

\makeatother
\newcommand*{\myexternaldocument}[1]{
    \externaldocument{#1}
    \addFileDependency{#1.tex}
    \addFileDependency{#1.aux}
}

\definecolor{cvprblue}{rgb}{0.21,0.49,0.74}
\usepackage[pagebackref,breaklinks,colorlinks,allcolors=cvprblue]{hyperref}
\usepackage[capitalize]{cleveref}
\crefname{section}{Sec.}{Secs.}
\crefname{table}{Table}{Tables}
\crefname{figure}{Fig.}{Figs.}

\ifarxiv \crefname{appendix}{App.}{Apps.}
\else \crefname{appendix}{Suppl.}{Suppls.} \fi

\unless\ifarxiv \myexternaldocument{_supplementary} \fi

\begin{document}
\title{\paperTitle}
\author{\authorBlock}
\maketitle

\begin{abstract}
Multi-hand semantic grasp generation aims to generate feasible and semantically appropriate grasp poses for different robotic hands based on natural language instructions. Although the task is highly valuable, due to the lack of multi-hand grasp datasets with fine-grained contact description between robotic hands and objects, it is still a long-standing difficult task. In this paper, we present Multi-GraspSet, the first large-scale multi-hand grasp dataset with automatically contact annotations. Based on Multi-GraspSet, we propose Multi-GraspLLM, a unified language-guided grasp generation framework, which leverages large language models (LLM) to handle variable-length sequences, generating grasp poses for diverse robotic hands in a single unified architecture. Multi-GraspLLM first aligns the encoded point cloud features and text features into a unified semantic space. It then generates grasp bin tokens that are subsequently converted into grasp pose for each robotic hand via hand-aware linear mapping. The experimental results demonstrate that our approach significantly outperforms existing methods in both real-world experiments and simulator. More information can be found on our project page \url{https://multi-graspllm.github.io}.
\end{abstract}
\section{Introduction}
\label{sec:intro}


Grasp generation~\cite{deligrasp,dexgraspnet,dexgraspnet2,DexGrasp-Diffusion,Grasp-as-You-Say,grasp-d,6-dof-graspnet,contact-graspnet,DGTR} is essential for robotic systems to interact with and manipulate their surroundings. Effective grasp generation is crucial for enhancing the versatility and adaptability of robotic systems, especially in complex scenarios where a wide range of tasks require precise handling. However, existing grasp generation methods mainly focus on predicting physically-stable grasp poses using specialized models for individual robotic hands, while overlooking grasp generation across different robotic hands and the role of semantics in grasp generation.

\begin{figure}[t]
  \centering
    \includegraphics[width=\linewidth]{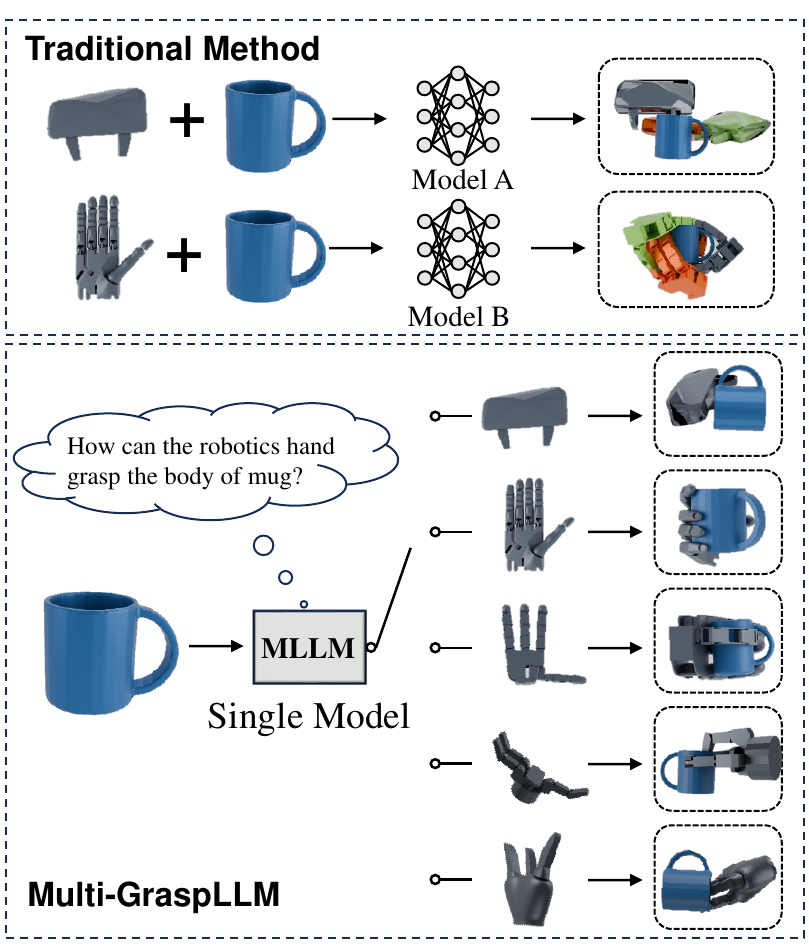}
   \caption{Multi-GraspLLM vs. traditional grasp generation method. Traditional grasp generation methods train separate models for each robotic hand, focusing primarily on generating physically stable grasps. In contrast, Multi-GraspLLM can use a single model to generate semantic-guided grasping poses adaptable to various robotic hands.}
   \vspace{-0.8cm}
   \label{overview}
\end{figure}

In this work, we focus on a challenging problem of \textbf{multi-hand semantic-guided grasp generation}. This task aims to generate feasible and semantically appropriate grasp poses for multiple robotic hands through natural language instructions. Unlike traditional approaches that require specific models for each individual robotic hand (Figure~\ref{overview}), this task aims to generate grasp poses of multiple hands with a single model. 
The unified approach not only offers greater flexibility by optimizing a single model when deploying, but also enhances generalization of grasp generation models through cross-hand learning. 
Furthermore, while existing methods lack semantic understanding in grasp generation, this task incorporates semantic guidance to generate semantically appropriate grasps based on textual descriptions.

Generally, achieving this goal involves two major challenges. The first challenge is to build a comprehensive dataset that includes various robotic hands and semantic descriptions. However, this is not a trivial task. Among existing multi-hand datasets~\cite{dexgraspnet,gendexgrasp,grasp-d,MultiGripperGrasp} that focus mainly on physically stable grasps, only one dataset~\cite{Grasp-as-You-Say} provides rich contact semantic information. However, its retargeting approach based on human grasp data~\cite{oakink} does not generalize well to robotic hands with different link and joint structures, especially multi-fingered systems with fewer than five fingers, such as the Jaco Hand~\cite{jaco} and Barrett Hand~\cite{barrett}.
Moreover, training a unified model for multiple robotic hands faces additional challenges when incorporating semantic guidance, as the same semantic instruction must be translated into different grasping behaviors for hands with distinctly different robotic structures. Most semantic-guided methods~\cite{Grasp-as-You-Say,human-like-grasp,semgrasp} train a separate model for each hand. Although DexGrasp-Diffusion~\cite{DexGrasp-Diffusion} based on DexGraspNet~\cite{dexgraspnet} trains a unified diffusion model for different robotic hands, it lacks semantic guidance to grasp objects. 

To address the first challenge, we propose Multi-GraspSet, the first large-scale multi-hand grasp dataset with contact annotations. Starting with pre-segmented objects~\cite{oakink,shapenet}, we first generated a large number of physically stable grasps using the unified grasp generation approach~\cite{dexgraspnet,contact-graspnet,graspit}. For each grasp case, we then added detailed contact annotations based on the segment annotation. Furthermore, we used LLM~\cite{gpt-4} to generate nearly 1.2M dialogue samples to support subsequent model training.

For the second challenge, based on the Multi-GraspSet, we design
Multi-GraspLLM to generate precise, feasible grasp poses for multiple
robotic hands. Multi-GraspLLM leverages the generalization capability and flexibility of large language models to handle variable-length inputs and outputs, enabling a single model to produce poses for different robotic hands.
The model takes a pair of point cloud and text instruction as input and outputs grasp angles for different robotic hands via a hand-aware linear mapping. For example, given the instruction \emph{``How can we use the Allegro Hand to grasp a glass by its rim?''}, Multi-GraspLLM can infer the optimal grasp angles for the Allegro Hand~\cite{allegro}. Similarly, when asked \emph{``How can we use the Shadow Hand to grasp a glass by its rim?''}, the same model can infer the appropriate grasp angles for the Shadow Hand~\cite{shadow}.

Through extensive experiments and evaluation of our dataset, we demonstrate that Multi-GraspLLM can successfully generate accurate grasp poses across multiple types of robotic hands while maintaining precise control over individual finger positions. Quantitative results in simulator and real-world experiments show significant improvements in both pose accuracy and grasp quality compared to existing methods. 
The model exhibits robust generalization capabilities across different grasp configurations while maintaining semantic consistency with natural language instructions.  

Our key contributions can be summarized as follows.
\begin{enumerate}
    \item We build Multi-GraspSet, the first large-scale multi-hand grasp dataset with rich contact annotations. Our dataset could help fill a gap in robotic hand grasp pose generation, particularly in multi-hand and semantic-guided grasping.
    \item We propose Multi-GraspLLM, an LLM-based multi-hand semantic grasp generation method that enables cross-hand robotic grasp generation while ensuring semantically proper grasps. Our work could also inspire related research of embodied foundation models for different robotics systems.   
    
    \item Experimental results show that Multi-GraspLLM achieves state-of-the-art performance in grasp generation. Moreover, joint training of a multi-hand model yields better results than training separate models for each hand.

\end{enumerate}
\section{Related Works}
\label{sec:Related_Works}
\subsection{Robotics Grasp Datasets}
Grasping datasets can generally be categorized into two types: physically stable and contact-aware. Most existing datasets focus on the former type. DexGraspNet~\cite{dexgraspnet,dexgraspnet2} and GenDexGrasp~\cite{gendexgrasp} use force-closure~\cite{force-closure} as a metric for simulator optimization. Acronym \cite{Acronym} and MultiGripperGrasp \cite{MultiGripperGrasp} using the physical simulator~\cite{issac} to filter out unstable grasps also support multiple robotic hands but have fewer objects. In contrast, contact-aware datasets transfer human hand poses to robotic hands. DexGYSNet~\cite{Grasp-as-You-Say} and Human-like-grasp~\cite{human-like-grasp} retarget the hand pose from the OakInk~\cite{oakink} dataset to the dexterous robotic hand and its own collected data with segmentation annotations. However, due to the difference in joints and links, DexGYSNet~\cite{Grasp-as-You-Say} and Human-like-grasp~\cite{human-like-grasp} typically only support robotic hands resembling human hands.
\begin{figure*}[ht]
  \centering
    \includegraphics[width=\linewidth]{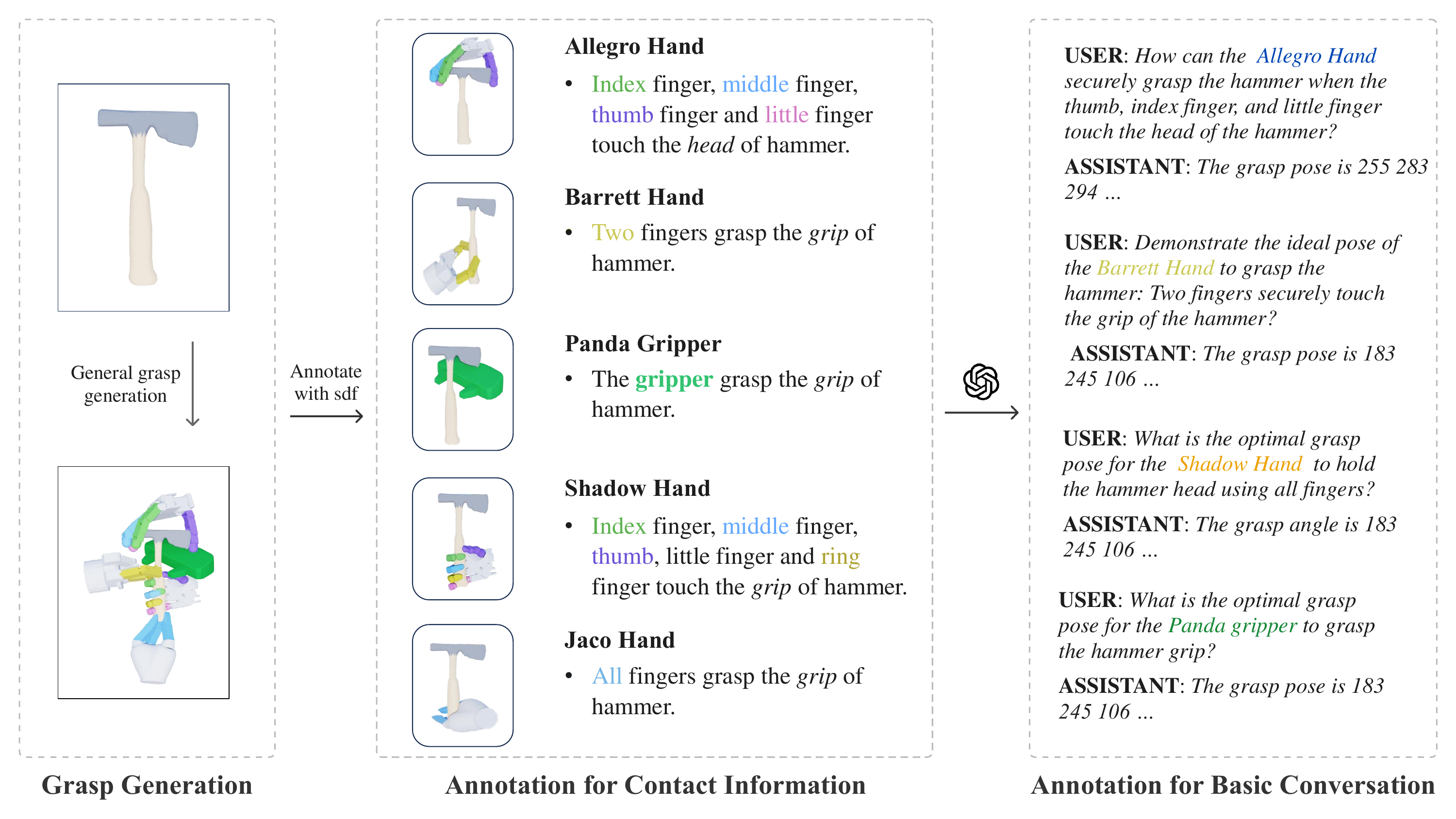}
   \caption{
Multi-GraspSet Construction Process. The initial unified grasp generation produces physically stable grasps. Then, through two levels of annotations, we generate pair data containing basic conversation with corresponding grasp pose for each robotics hand.
   }
    \vspace{-0.5cm}
   \label{vis_dataset}
\end{figure*}

\subsection{Language-guided Grasp Generation}
Although extensive research has been conducted on physically-stable grasp generation methods~\cite{dexgraspnet,dexgraspnet2,grasp-d,fast-grasp-d,GraspTTA,DGTR,force-closure,contact-graspnet,6-dof-graspnet}, there remains a significant gap in studies focused on language-guided functional grasp generation. The aim is to generate an appropriate grasp pose on a specific functional part of an object based on the description of the grasping scenario.
LERF-TOGO~\cite{Lerf-togo} and GraspSplats~\cite{graspsplats} applied NERF~\cite{nerf} and 3DGS~\cite{3dgs} to model 3D semantics scenes, allowing the gripper to grasp the specific object part based on language queries. Human-like-grasp~\cite{human-like-grasp} trained a CVAE~\cite{cvae} model that allowed the generation of different functional grasping. However, it cannot support an open vocabulary. DexGYS~\cite{Grasp-as-You-Say} addresses this limitation by training a language-conditioned diffusion model to generate dexterous grasp poses based on language descriptions. At the same time, large models greatly contribute to language-conditioned grasp generation. RealDex~\cite{realdex} leverages Gemini~\cite{gemini} to score the generated grasps, filtering out those that satisfy the language description. SemGrasp~\cite{semgrasp} further extended this idea. It fine-tuned a large language model and used VQ-VAE~\cite{VQ-VAE} to discretize low-dimensional hand pose parameters. However, all of these methods only support a single type of hand.  

\subsection{Multimodal LLMs in Robotics}
Multimodal LLMs play a crucial role in reasoning for embodied intelligence, especially with the full development of models such as LLaVA~\cite{llava} and LLaMA~\cite{llama}. Several models~\cite{affordancellm,deligrasp,Reasoning-grasping,Reasoning-Tuning-Grasp} leverage the reasoning capabilities of vision-language models~\cite{llava} to identify grasping regions in scene images. Vision-language-action model~\cite{open-vla,Open-x,rt-2, RoboFlamingo, ADriver-I,Vision-language-foundation,robomatrix}, built on the LLM architecture, enables the generation of discrete actions based on both visual and language input from a scene. Additionally, 3D large language models~\cite{shapegpt,meshgpt,minigpt-3d,pointllm} extend the structure of LLaVA by exchanging the 2D visual encoders with various 3D point cloud encoders such as PointNet~\cite{pointnet,pointnet++} or PointBERT~\cite{pointbert}, which enable effective extraction of geometric features from unordered point sets and scene understanding. Methods like~\cite{embodied,3d-vla,affordance-agent,LL3DA} refine tasks in embodied environments, while SemGrasp~\cite{semgrasp} focuses on more specific tasks, displaying human hand grasp poses based on inputs from object point cloud.



\section{Multi-GraspSet Dataset}
\label{sec:DexContactSet}

To our best knowledge, we proposed the first large-scale multi-hand grasp dataset with semantic contact annotation, namely Multi-GraspSet. This dataset was created through unified grasp generation and LLM assistance system. We detail the basic statistics in Sec.~\ref{sec:DexContactSet_statics}, and then introduce the dataset construction process in Sec.~\ref{sec:DexContactSet_construction}.

\subsection{Dataset Statistics}
\label{sec:DexContactSet_statics}
\begin{table}[h!]
\begin{center}
\begin{tabular}{l|c|c|c|c}
\toprule
Dataset & Hand  &  Object & Grasp  & Con. \\ 
\midrule
 DexGYSNet \cite{Grasp-as-You-Say} &1 &1800 & 50k & -\\
 CapGrasp \cite{semgrasp} &1 & 1800 & 50k & 280k\\
 Multi-GraspSet &\textbf{5} & \textbf{2100}  & \textbf{140k} & \textbf{1.1M} \\
\bottomrule
\end{tabular}
\end{center}
\vspace{-0.2cm}
\caption{Comparison of the contact-aware grasp dataset. ``Grasp'' and ``Con.'' stand for the number of grasp pose and conversation.}
\label{table_dataset}
\vspace{-0.2cm}
\end{table}

\begin{figure}[h!]
    \includegraphics[width=\linewidth]{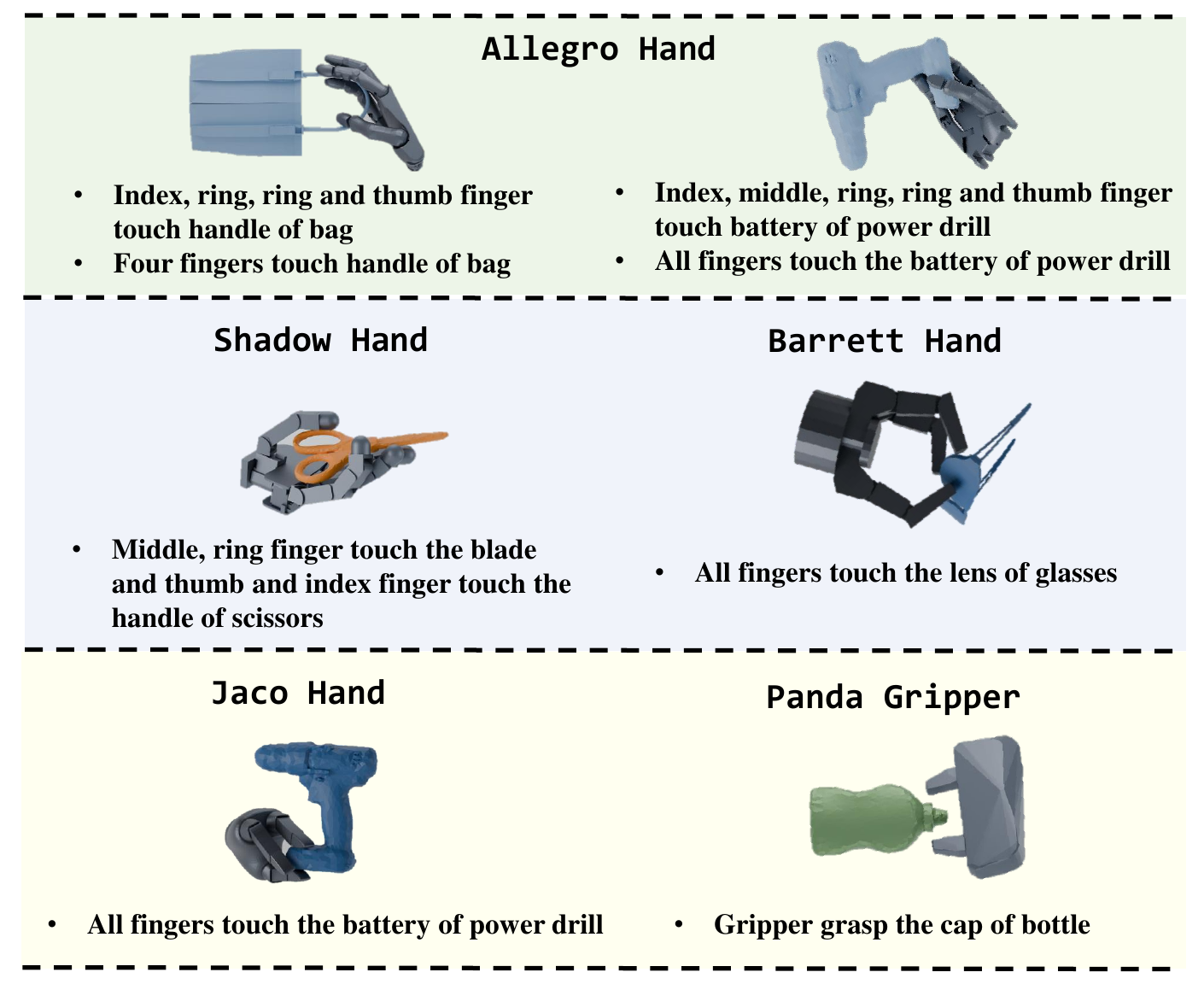}
   \caption{Illustration of Multi-GraspSet. Our dataset includes grasp poses of common objects across five robotic hands with two types of contact annotations.}
   
   \label{vis_dataset_grasp}
    \vspace{-0.5cm}
\end{figure}

Multi-GraspSet is a large-scale dataset that contains the grasp of five popular robotic hands, 2.1k object point clouds, 140k grasp pose pairs, and more than 1M conversations (Table~\ref{table_dataset}). The dataset features two main types of annotations: \textbf{contact information} and \textbf{basic conversation}. For contact information, we provide precise object contact annotations for each finger, and when all contact-involved fingers interact with the same object part, we provide a more general annotation emphasizing the number of fingers participating in the grasp, as illustrated in Figure~\ref{vis_dataset_grasp}. Moreover, to enable Multi-GraspLLM to generate grasps following natural language instructions, we construct basic conversations for each grasp pose with different control levels, comprising 5-10 diverse question-answer hand-aware basic conversations focusing on the grasp pose (Figure~\ref{vis_graspllm_grasp}).

\subsection{Dataset Construction}
\label{sec:DexContactSet_construction}
We first apply the unified grasp generation methods~\cite{dexgraspnet,contact-graspnet} to generate physically stable grasp poses of collected meshes for dexterous hands and grippers. Then we design a
contact information annotation method using signed distance fields (SDFs), which can produce fine-grained contact information between the finger and the object, accurately capturing the interaction between the gripper and the object. Finally, with the assistance of large language models \cite{gpt-4}, we generate a significant amount of conversational data, facilitating the dialogue-level understanding of robotic grasping.

\subsubsection{Unified Grasp Generation}
\label{Unified}
We generate grasp poses of five robotic hands through the unified grasp generation methods~\cite{dexgraspnet,graspit,contact-graspnet} (Figure~\ref{vis_dataset}). Specifically, DexGraspNet~\cite{dexgraspnet} and GraspIt~\cite{graspit} proposes an analysis-based method for dexterous hands, working without learning. Contact-GraspNet~\cite{contact-graspnet} introduces an end-to-end grasp generation network for grippers with strong generalization. 
We first collect 2,100 meshes from OakInk~\cite{oakink} and ShapeNet~\cite{shapenet}, applying convex decomposition to ensure that the meshes are watertight. Since we aim at multi-hand grasp generation, we then use DexGraspNet~\cite{dexgraspnet} and GraspIt~\cite{graspit} to generate grasp poses for dexterous robotic hands (Shadow Hand~\cite{shadow}, Allegro Hand~\cite{allegro}, Barrett Hand~\cite{barrett} and Jaco hand~\cite{jaco}), and Contact-GraspNet~\cite{contact-graspnet} for grippers (Panda Gripper~\cite{panda}). 

\subsubsection{Annotation for Contact Information}
\label{Contact}
Once the multi-hand grasp generation is complete, we use the Signed Distance Function (SDF) to annotate the contact information shown in Figure~\ref{vis_dataset}. Specifically, we assign a category label to each mesh face based on the segmentation annotations from the original dataset. For each object, we perform uniform sampling to generate a set of points on the object's surface. Next, we use Kaolin~\cite{kaolin} to compute the SDF values from these points to the different links of the robotic hand.
When the SDF value falls below a predefined threshold $\epsilon$, we consider the corresponding link to be in contact with the object:
\begin{equation}
C(link_i)= \mathbb{I}(SDF(P_{link_i}|O) <  \epsilon)
\label{eq1}
\end{equation}
where $C(link_i)$ indicates whether the $i$-th link is in contact with the object, $O$ represents the object's mesh, and $P$ denotes the point sampled from the robotic hand.
Additionally, based on the category labels of object faces, we annotate which specific part of the object the link is in contact with. As briefly mentioned in Sec.~\ref{sec:DexContactSet_statics}, we use two types of annotation to describe the finger contacts: a detailed annotation that specifies individual fingers—``\emph{Index, middle, thumb, and ring fingers contact the hammer's grip}'' and a general type for cases where all fingers contact the same part—``\emph{Four fingers grasp the grip of the hammer}.''  
 The general annotation is used to improve text quality and readability when fingers share the same contact location on the object. 

\subsubsection{Annotation for Basic Conversation}
\label{Conversation}
Inspired by SemGrasp~\cite{semgrasp}, we develop an LLM-assisted language guidance annotation system~\cite{gpt-4} to construct basic conversational datasets for robotic grasp tasks. We begin by prompting the LLM to generate templates that incorporate object names, robotic hand types, grasp parts, and contact details. 

We categorize basic conversation into three types: low-level, middle-level, and high-level instructions, with levels of detail ranging from coarse to fine. 
Low-level instructions do not provide contact information and include only basic questions (e.g., ``How do you grasp the \{object\} using the \{hand type\}?'')
Middle-level instructions specify the part to be grasped (e.g., ``How do you grasp the \{part\} of the \{object\} using the \{hand type\}?'')
High-level instructions include all relevant information (e.g., ``Demonstrate the ideal pose of the \{hand type\} to grasp the \{object\}: \{contact info\}.'')
However, due to the complexity of our contact information compared to CapGrasp~\cite{semgrasp}, fixed templates often lead to awkward phrasing. To address this, we use GPT-4o~\cite{gpt-4} to refine each sentence, ensuring varied and natural language throughout. More details about our annotation system can be found in the Supplementary Material.

\begin{figure*}[ht]
  \centering  
   \includegraphics[width=\linewidth]{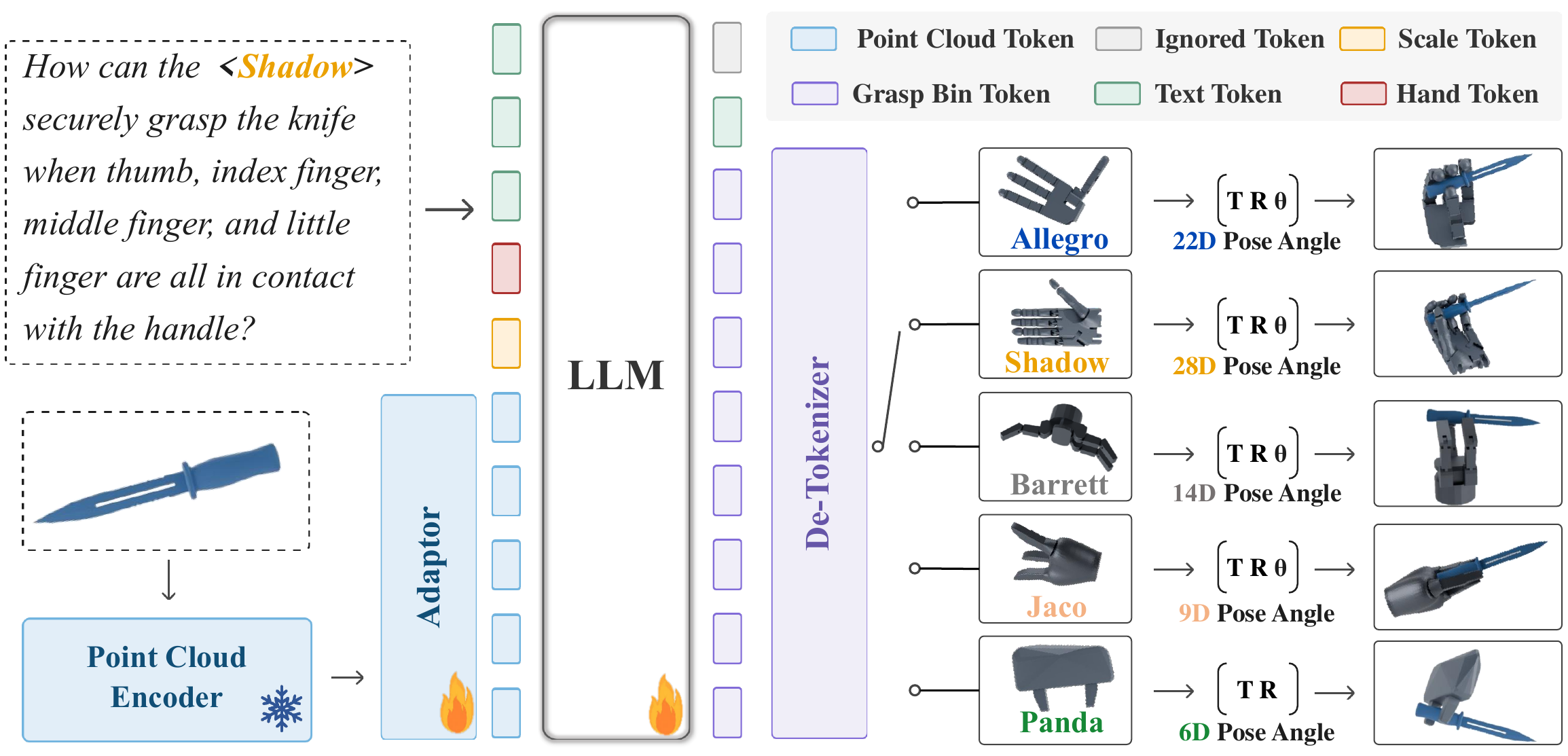}
   \caption{Multi-GraspLLM model. The point encoder extracts point clouds from objects and maps them with language descriptions into the same latent space. The LLM backbone then generate grasp bin tokens as output. Finally, we convert these grasp bin tokens into corresponding grasp angles for each robotic hand.}
   \label{pipeline}
\vspace{-0.5cm}
\end{figure*}

\section{Method}
\label{sec:LLM}
We propose Multi-GraspLLM, a framework based on a large language model~\cite{vicuna} that generates appropriate grasp poses for different robotic hands by integrating point cloud data with natural language descriptions. We describe the detail of our framework in Sec.~\ref{sec:LLM:GraspLLM}, then we show the data format and training strategy in Sec.~\ref{sec:LLM:Training_data} and Sec.~\ref{sec:LLM:Strategy}, respectively.
\subsection{Multi-GraspLLM}
\label{sec:LLM:GraspLLM}

We initially apply a hand-aware discretization approach for grasp angles~\cite{rt-2,open-vla}. These discretized grasp bins are then used to train the Multi-GraspLLM, enabling it to generate appropriate grasp poses across multiple robotic hands as shown in Figure~\ref{pipeline}.

\vspace{1mm}
\noindent\textbf{Multi-hand Discretization.} 
To transform the prediction of continuous grasp angle into a more tractable classification task for language models, we convert continuous angles into discrete tokens. Specifically, we uniformly discretize the grasp angles by dividing the valid range between the hand-specific angle lower bound $L_{hand}$ and the upper bound $U_{hand}$ into $N$ bins, where both hand-specific bounds are computed across the entire dataset. The bin width $W_{hand}$ is defined as $W_{hand}=(U_{hand}-L_{hand})/N$. Then we map a continuous angle $p$ to its corresponding discrete bin $B$:
\begin{equation}
B=\frac{p-L_{hand}}{W_{hand}}, hand\in \{Allegro,Shadow,\cdots \}
\label{eq2}
\end{equation}
This discretization scheme is independently determined for each robotic hand based on its grasping dataset, ensuring efficient prediction while maintaining precision.

\vspace{1mm}
\noindent\textbf{Grasp Pose Generation.}
Multi-GraspLLM comprises three components: 1) a 3D feature encoder $f_{pc}$ that extract objects geometry information; 2) a modality adaptor $f_{m}$ that aligns object geometry with language features; 3) a large language model $m_{llm}$ to interpret natural language grasp instructions and generate the grasp pose based on the 3D geometry of object.

Given an object point cloud $P\in \mathbb{R}^{8192}$ and a language description $T$, the point cloud encoder first encodes $P$ into a sequence of feature tokens $s=f_{pc}(P)$, where $s\in \mathbb{R}^{512\times384}$. The modality adaptor then aligns these point cloud features into the language space $T_p=f_{m}(s)$, where $T_p\in \mathbb{R}^{512\times4096}$, to match the language features $T_l$ encoded by the LLM language encoder. These two token sequences are concatenated to form a mixed token sequence $T_m=[T_p,T_l]$, which serves as input to the LLM backbone. 
The LLM backbone processes this sequence auto-regressively, generating outputs $O_i=m_{llm}(O_1,...,O_{i-1},T_m)$ for the sequence $[O_1,O_2,...,O_n]$. Finally, the de-tokenizer converts the output sequence into a grasp bin $B=[B_1,B_2,...]$. Each bin acts as a special token, analogous to a word in a large language model. Each bin serves as a special token, analogous to the output of words in LLM. Once the actions are processed into a sequence of bins, 
Multi-GraspLLM is trained with a standard next-token prediction objective, evaluating the Cross Entropy loss on the predicted grasp bin tokens and generated natural language to enhance the model's unified generalization capability.

We then de-discretize the grasp bin to grasp angle of specific robotic hand. Given the predicted grasp bin $B=[B_1,B_2,...]$ from the de-tokenizer, the discrete bin index is converted linearly back to a continuous value as follows: 
\begin{equation}
[T,R,\theta]= L_{hand} + B_iW_{hand}
\label{eq3}
\end{equation}
where $[T;R] \in \mathbb{R}^6$ represents the 6-D pose of the robotic hand's wrist, and $\theta\in \mathbb{R}^{d}$ denotes the joint angles specific to different robotic hands (e.g , $d=22$ for Allegro Hand~\cite{allegro} as shown in the Fig.~\ref{pipeline}). The detailed implementation is provided in the supplementary material.

\subsection{Data Format} \label{sec:LLM:Training_data}

\begin{figure}[h!]
    \includegraphics[width=\linewidth]{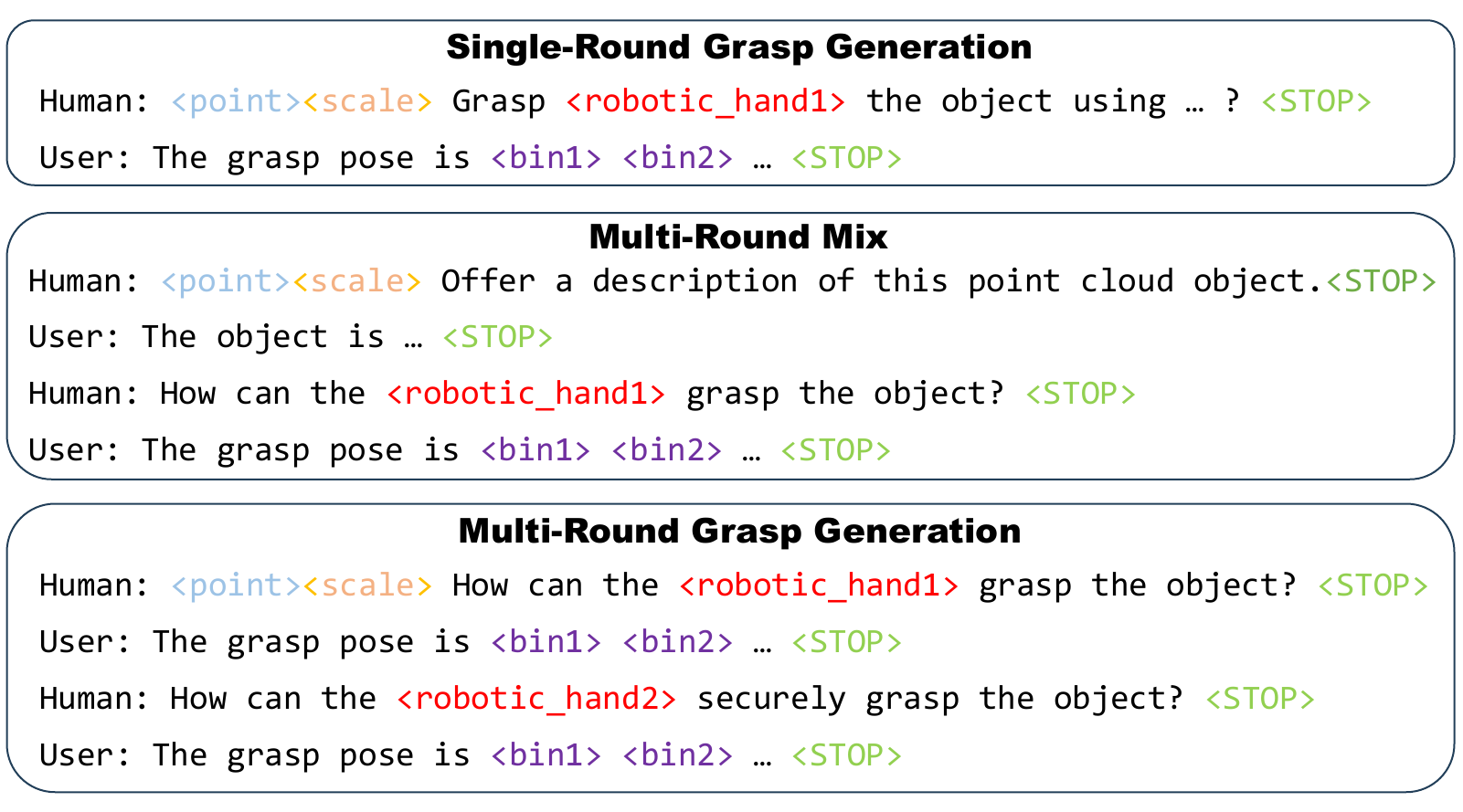}
   \caption{The template of training data used to train Multi-GraspLLM.}
   \label{train_data}
\vspace{-0.3cm}
\end{figure}

As shown in Figure~\ref{train_data}, we introduce three special tokens in the Vicuna~\cite{vicuna} backbone fine-tuning: 1) Robotics hand special token for hand identification, 2) Scale special token for point cloud size normalization, and 3)  Grasp special bin token mapping to 512 discretization bins. These tokens work together to ensure accurate information processing during both tokenization and de-tokenization.

With these specialized tokens, we combined and refined the basic conversations from Multi-GraspSet and the caption task. The final dataset consists of four types of conversation, as shown in Figure~\ref{train_data}: 1) \textbf{Single-Round Grasp Generation}, 2) \textbf{Multi-Round Mix}, and 3) \textbf{Multi-Round Grasp Generation}. The first type involves single-round interactions, where the model predicts grasping angles based on simple queries like ``\emph{How can the \{robotic hand\} grasp the object?}''
The latter two types feature multi-round interactions. ``Multi-Round Mix'' starts with object class description and follows with grasp prediction, and  ``Multi-Round Grasp Generation'' supports more diverse dialogues involving different hands and requiring sophisticated grasp predictions, going beyond the basic approach.
We purposely incorporated captioning tasks to enhance the model's 3D geometric understanding and reduce its reliance on text-only inputs. Our ablation studies in Sec.~\ref{sec:ablation} validate that this strategy significantly improves performance, achieving a balanced interpretation of spatial geometry and grasping actions.

\subsection{Training Strategy} \label{sec:LLM:Strategy}

To effectively train our model, we employ a two-stage training procedure consisting of multimodal alignment and instruction tuning.

\vspace{1mm}
\noindent\textbf{Stage 1: Multimodal Alignment.}
In multimodal alignment, we aim to align the point cloud embedding space as closely as possible with the language feature space through the adaptor. 
We freeze the LLM backbone and point cloud encoder while fine-tuning only the modality adaptor. The fine-tuning process incorporates single-round conversations ``Single-Round Grasp Generation'' described in Sec.~\ref{sec:LLM:Training_data}.

\vspace{1mm}
\noindent\textbf{Stage 2: Instruction Tuning.}
In instruction tuning, we want our LLM to thoroughly understand the grasp generation task. We always keep the point cloud encoder weights frozen, and continue to update both the pre-trained weights of the adaptor and LLM in Multi-GraspLLM. In this stage, we tend to use complex conversations that consist of all types as described in Sec.~\ref{sec:LLM:Training_data}.

\section{Experiments}
\begin{table*}[th]
\setlength{\tabcolsep}{0.7pt}

\begin{center}
\begin{tabular}{l|c c c|c c c|c c c|c c c|c c c|c c c}
\toprule
\multirow{2}{*}{Model} & \multicolumn{3}{c|}{Allegro} & \multicolumn{3}{c|}{Shadow}& \multicolumn{3}{c|}{Barrett}& \multicolumn{3}{c|}{Jaco} & \multicolumn{3}{c|}{Panda} & \multicolumn{3}{c}{Avg} \\

& CD$\downarrow$ & Pen$\downarrow$ & Suc $\uparrow$ & CD $\downarrow$& Pen$\downarrow$ & Suc $\uparrow$ & CD$\downarrow$ & Pen$\downarrow$ & Suc$\uparrow$ & CD$\downarrow$ & Pen $\downarrow$& Suc $\uparrow$& CD$\downarrow$ & Pen $\downarrow$& Suc $\uparrow$& CD$\downarrow$ & Pen $\downarrow$& Suc $\uparrow$ \\
\midrule
\midrule
GroundTruth & 0.65 & 1.08 & 0.29 & 1.07 & 0.79 & 0.38 & 1.09 & 0.43 & 0.31 & 1.38 & 0.59 &0.37 & 0.65& 0.41& 0.44 & 0.97 &0.66&0.36 \\
DexGraspNet~\cite{dexgraspnet} & 0.62 & 1.27 & 0.31 & 0.94 & 0.83 & 0.35 & 1.30 & 0.63 & 0.21 & 1.20 & 0.45 &0.22 & /& /& / &1.02&0.80&0.28  \\
SceneDiffuser~\cite{Scene-diffusers} & 0.74 & 1.04 & 0.31 & 1.02 &  0.86 & 0.34 & 1.21 &0.72 & 0.24 & 1.58 & 0.71 &0.29 & / &/ & / &1.14 & 0.76 &0.30 \\
GraspNet~\cite{contact-graspnet} & / & / & / &/  & /& /&/ & / &/ &/ & / & /  & 0.62& 0.43 &0.46  & / &/ & /\\
DexGYS~\cite{Grasp-as-You-Say} & 0.5 & 1.18 & 0.27 & 0.54 & 0.83 & 0.39 & 0.78 &0.29& 0.29 &0.67 & 0.43& 0.35 & / &/ & / & 0.62 & 0.68 &0.32 \\

\rowcolor[gray]{.9} Multi-GraspLLM-split & 0.41&	1.04	&0.32	&0.48	&0.94	&0.41	&0.56&	0.38	&0.31	&0.45&	0.68	&0.38&	0.25&	\textbf{0.40}&	0.45&	0.43&	0.69	& 0.37  \\

\rowcolor[gray]{.9} Multi-GraspLLM-mix & \textbf{0.37} &	\textbf{1.00}	& \textbf{0.34}	&\textbf{0.42} &	\textbf{0.75}	&\textbf{0.43}	&\textbf{0.48}	& \textbf{0.27}	&\textbf{0.33}	& \textbf{0.34} &	\textbf{0.41}	&\textbf{0.43}	& \textbf{0.20} &	\textbf{0.40}	& \textbf{0.48} &	\textbf{0.36}&	\textbf{0.57}	& \textbf{0.40} \\
\toprule
\end{tabular}
\caption{Comparison of SOTA methods and Multi-GraspLLM. GroundTruth stands for the preformance of the raw training dataset. ``-mix'' denotes models trained on mixed datasets, while ``-split'' indicates models trained separately for each robotic hand. ``Avg'' stands for the average performance of all hands.}
\vspace{-0.9cm}
\label{table_main_result}
\end{center}
\end{table*}

   

\subsection{Dataset and Evaluation Metrics}
Our Multi-GraspSet dataset is split into three parts: 80\% for training, 10\% for validation, and 10\% for testing. The point clouds in the test set are completely separate from those used in training and validation. 

Multi-GraspLLM is evaluated in two key aspects: grasp intention and physical stability.
To evaluate grasp intention accuracy, we employ \textbf{Chamfer distance (CD)} (cm), which quantifies the spatial alignment between the predicted and ground truth robotics hand point clouds. This metric measures how well our inferred grasps match the intended grasp locations on objects, where lower values indicate better alignment.
To evaluate grasp quality, we employ three metrics: the \textbf{maximum penetration distance (Pen)} (cm) between the hand and target object measures physical feasibility, the \textbf{grasp success rate (Suc)} in Isaac Sim. More details are in our supplementary material.
\subsection{Implementation Details}
\label{Implementation_Details}
For constructing Multi-GraspSet, we employ DexGraspNet~\cite{dexgraspnet}, GraspIt~\cite{graspit} and Contact-GraspNet~\cite{contact-graspnet} as our grasp generation methods. For DexGraspNet~\cite{dexgraspnet}, we generate 128 grasp poses for the Allegro Hand~\cite{allegro} and 160 for the Shadow Hand~\cite{shadow} per object, using Adam optimizer with 6000 epochs. For GraspIt~\cite{graspit}, we generate 30 grasps for Jaco hand~\cite{jaco} and 60 grasps for Barrett hand~\cite{barrett}. For Contact-GraspNet~\cite{contact-graspnet}, we quickly sample 256 points from each point cloud and generate 200 grasps. 


During training, we perform multimodal alignment for 3 epochs with a learning rate of 2e-3, followed by instruction tuning for another 3 epochs with a learning rate of 2e-5. For each object, we uniformly sample 8,192 points from its mesh to generate the point cloud. The grasp angles are discretized into 384 bins, with their upper and lower bounds for different robotic hands determined by their entire dataset.
\subsection{Comparison with State-of-the-Art Methods}
We compared our Multi-GraspLLM method with existing open-source SOTA approaches. Since there is currently no open-source method for multi-dexterous hand and gripper grasping generation, we conducted separate comparisons on corresponding end-effectors: SceneDiffuser~\cite{Scene-diffusers}, DexGYS~\cite{Grasp-as-You-Say} and DexGraspNet~\cite{dexgraspnet} were evaluated on dexterous hands, both trained on our dataset, while Contact-GraspNet~\cite{contact-graspnet} was evaluated on parallel grippers. The main results are shown in Table~\ref{table_main_result}, ``Multi-GraspLLM-split'' means that our Multi-GraspLLM model is trained separately on the dataset for each manipulator, while ``-mix'' refers to mixed training. The results show that mixed training outperforms separate training by a large margin, indicating that joint training can significantly improve model performance. Our methods trained separately or mixed outperform all existing baselines, especially in intention accuracy, with better physical stability and less penetration. Unlike DexGraspNet~\cite{dexgraspnet}, which adds specific loss functions for physical stability, and the two-stage method DexGYS~\cite{Grasp-as-You-Say}, our approach is end-to-end without any extra loss functions. Our approach outperforms all existing methods and surpasses our original Multi-GraspSet dataset, as demonstrated by the ``GroundTruth'' presented in Table~\ref{table_main_result}.

\begin{figure*}[ht]
    \includegraphics[width=\linewidth]{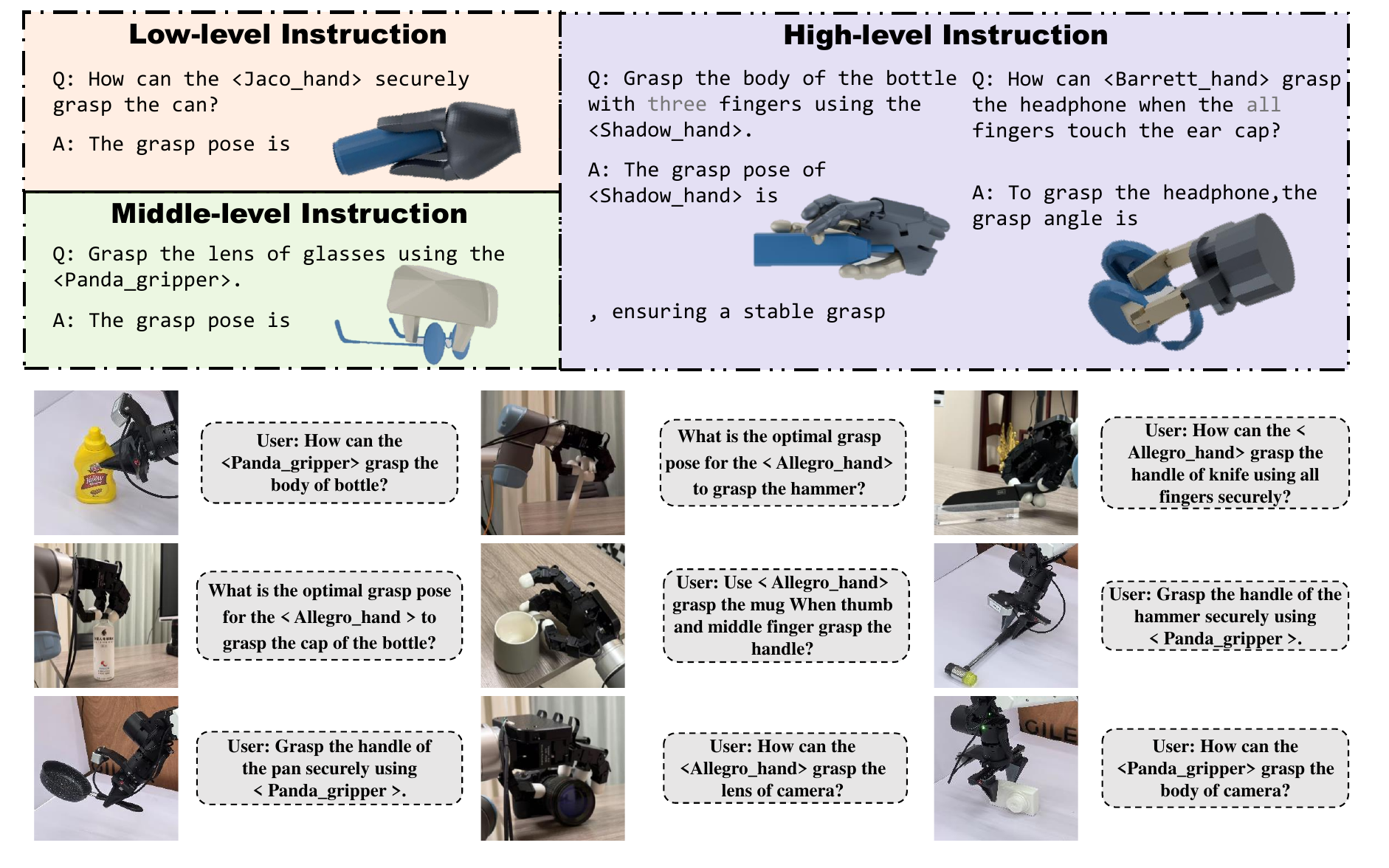}
    
   \caption{Visualization of the grasp pose generated by Multi-GraspLLM.}
   \label{vis_graspllm_grasp}
   \vspace{-0.3cm}
\end{figure*}

Figure~\ref{vis_graspllm_grasp} shows that Multi-GraspLLM can generate grasp poses based on different contact information: the low-level yields grasps without contact data, the mid-level uses object part information and the high-level leverages finger contact for fine control.

\subsection{Ablation Study}
\label{sec:ablation}
To verify the effectiveness of each component of our method, we performed ablation studies using the average performance of all hands, investigating the impact of each special token and the presence of two-stage training. We also explored the effect of grasp bin length and investigated how the structure of training data affects the model's performance.


\begin{table}[h]
\setlength{\tabcolsep}{3pt}

\begin{center}
\begin{tabular}{l|c|c|c}
\toprule
Method & CD$\downarrow$ & Pen$\downarrow$ & Suc$\uparrow$  \\
\midrule
Multi-GraspLLM & \textbf{0.36}  & \textbf{0.57} &  \textbf{0.40} \\
w/o hand token &0.38 & 0.63  & 0.38 \\
w/o scale token &0.42 & 0.71  & 0.35 \\
w/o grasp token &0.41 & 0.62  & 0.36  \\
w/o 2-stage training & 0.48  & 0.74 & 0.32 \\

\midrule
\end{tabular}
\caption{Ablation study of our Multi-GraspLLM grasp generation.
 }
 \vspace{-0.3cm}
\label{table_ablation1}
\end{center}
\end{table}
\vspace{-0.3cm}


We first performed ablation on each special token and two-stage training. As shown in Table~\ref{table_ablation1}, the two-stage training and the grasp bin token significantly improve performance. A single-stage model struggles to respond effectively to newly introduced grasp bin tokens never encountered before. Due to the incomplete encoding vocabulary for numbers in Vicuna's~\cite{vicuna} tokenizer, larger numbers without grasp bin token such as ``400'' may be decoded as separate tokens [``40'',``0''], which further hinders the model's grasp bin comprehension. Additionally, 
since PointBERT~\cite{pointbert} normalizes point clouds and thus missing overall scale information,
a scale token helps the model better understand the shape of the point clouds. Lastly, the robotic hand special token still exerts a significant impact, though we only use five types of robotic hands, which the model can memorize by name. 

\begin{table}[h]
\setlength{\tabcolsep}{3pt}

\begin{center}
\begin{tabular}{l|c|c |c c }
\toprule
 Number& CD$\downarrow$ & Pen$\downarrow$ & Suc $\uparrow$  \\
\midrule
512bin&  0.38 &  0.62 & \textbf{0.41}  \\
384bin& \textbf{0.36}  & \textbf{0.57} &  0.40  \\
256bin& 0.40 & 0.61  & 0.36 \\
\midrule
\end{tabular}
\caption{Effect of grasp bin number. }
\label{table_ablation2}
\vspace{-0.9cm}
\end{center}
\end{table}

Next, we investigated the effect of the number of grasp bins. The choice of bin number presents a trade-off: larger bin numbers (resulting in smaller bin widths) provide higher resolution, but make prediction more challenging. Smaller bin numbers simplify prediction but reduce the precision of grasp generation. We therefore selected a moderate bin width to balance these competing factors. Moreover, our method with all these bin number configurations consistently outperforms the existing methods (Table~\ref{table_main_result}).

\begin{table}[h]
\setlength{\tabcolsep}{1.5pt}
\centering

\begin{tabular}{c c c | c c c c}
\toprule
Single-Grasp & Mix & Multi-Grasp & CD$\downarrow$ & Pen$\downarrow$ & Suc$\uparrow $ \\ 
\midrule
\cmark &  &  & 0.40 & 0.67 & 0.37  \\ 
 \cmark& \cmark &  & 0.39 & 0.62 & 0.35  \\ 
\cmark & \cmark & \cmark & \textbf{0.36} & \textbf{0.57} & \textbf{0.40}  \\ 
 \bottomrule
\end{tabular}
\caption{Ablation on training data for instruction tuning.}
\label{table_ablation3}
 \vspace{-0.3cm}
\end{table}

We then performed an ablation study on the type of data for instruction tuning. As described in Sec.~\ref{sec:LLM:Training_data}, our instruction tuning data consists of three types,``Single-Grasp'':``Single-Round Grasp Generation'', ``Mix'': ``Multi-Round Mix'' and ``Multi-Grasp'': ``Multi-Round Grasp Generation''. As shown in Table~\ref{table_ablation3}, fine-tuned on data with more complex and multi-round conversations from ``Single-Grasp'',``Mix'' to   ``Multi-Grasp'', our model can achieve consistently better performance. Conceptually, training data with more complex structure allows the model to gain better performance through multitask learning. The introduction of auxiliary tasks (i.e. caption task) prevents the network from exclusively focusing on the primary task, which reduces the risk of overfitting. This training strategy also ensures that the model does not excessively concentrate on the processing of input language while neglecting geometric features of input point clouds, ultimately leading to improved grasp prediction.

\section{Conclusion}
In this work, we present the first multi-hand grasp dataset with rich semantic guidance. Leveraging this dataset, we propose Multi-GraspLLM, a multimodal LLM method for multi-hand grasp generation. Our method can generate grasp poses for different robotic hands based on linguistic grasp descriptions. Our approach fills a gap in the cross-hand semantic grasp generation field. In the future, we plan to collect larger data using more types of robotic hands and objects, which can further enhance the ability of Multi-GraspLLM to generalize to real-world scenarios, ultimately driving advancements in multi-robot collaboration and human-robot interaction.

{\small
\bibliographystyle{ieeenat_fullname}
\bibliography{main}
}
\appendix 
\clearpage

In this supplementary material, we first introduce the implementation details of our method (Sec.~\ref{Implementation}). Then we show additional ablation experimental (Sec.~\ref{Ablation}) and real world experimental results (Sec.~\ref{Real}). 

\section{Additional Implementation Details}
\label{Implementation}

\subsection{Dataset Annotation}
As described in our main paper, we used GPT-4o~\cite{gpt-4} to generate our basic conversation dataset. Specifically, we developed an LLM-assisted annotation method, which is based on well-crafted instruction prompts to generate natural and diverse conversations. In the following, we provide details on our carefully designed system prompt.

To generate diverse conversations based on different grasp contact information for each object, we adopted a template-based approach. We first used LLM~\cite{gpt-4} to generate rough templates, then filled these templates with the information of a specific grasp case, and finally used LLM~\cite{gpt-4} again to refine each dialogue, ensuring diverse and fluent conversations.

\begin{table}[h]
\fbox{\parbox{\linewidth}{
    \colorbox{gray!10}{
        \parbox{\dimexpr\linewidth-2\fboxsep\relax}{
            - USER: You now need to generate more instruction templates based on a series of templates that I provide to you. These templates focus on asking questions or giving instructions about grasping from the perspective of a robotic hand. Each new template should include ``object\_name'', ``contact\_info'', and ``robotics\_hand''.
            Below are the templates I provided to you. You need to create more templates similar to these, focusing on different grasping perspectives:
            
            templates = 
            [ f``How can the \{robotics\_hand\} securely grasp the \{object\_name\} when \{contact\_info\}?'', 
            
            f``What is the ideal grasping pose for the \{robotics\_hand\} when \{contact\_info\} with the \{object\_name\}?''
            ]
            
            The output format should be:
            templates = [...]
        }
    }
}}
    \caption{System prompt of generating initial rough template.}
    \label{initial_prompt}
\end{table}

\begin{figure*}[h]
    \includegraphics[width=\linewidth]{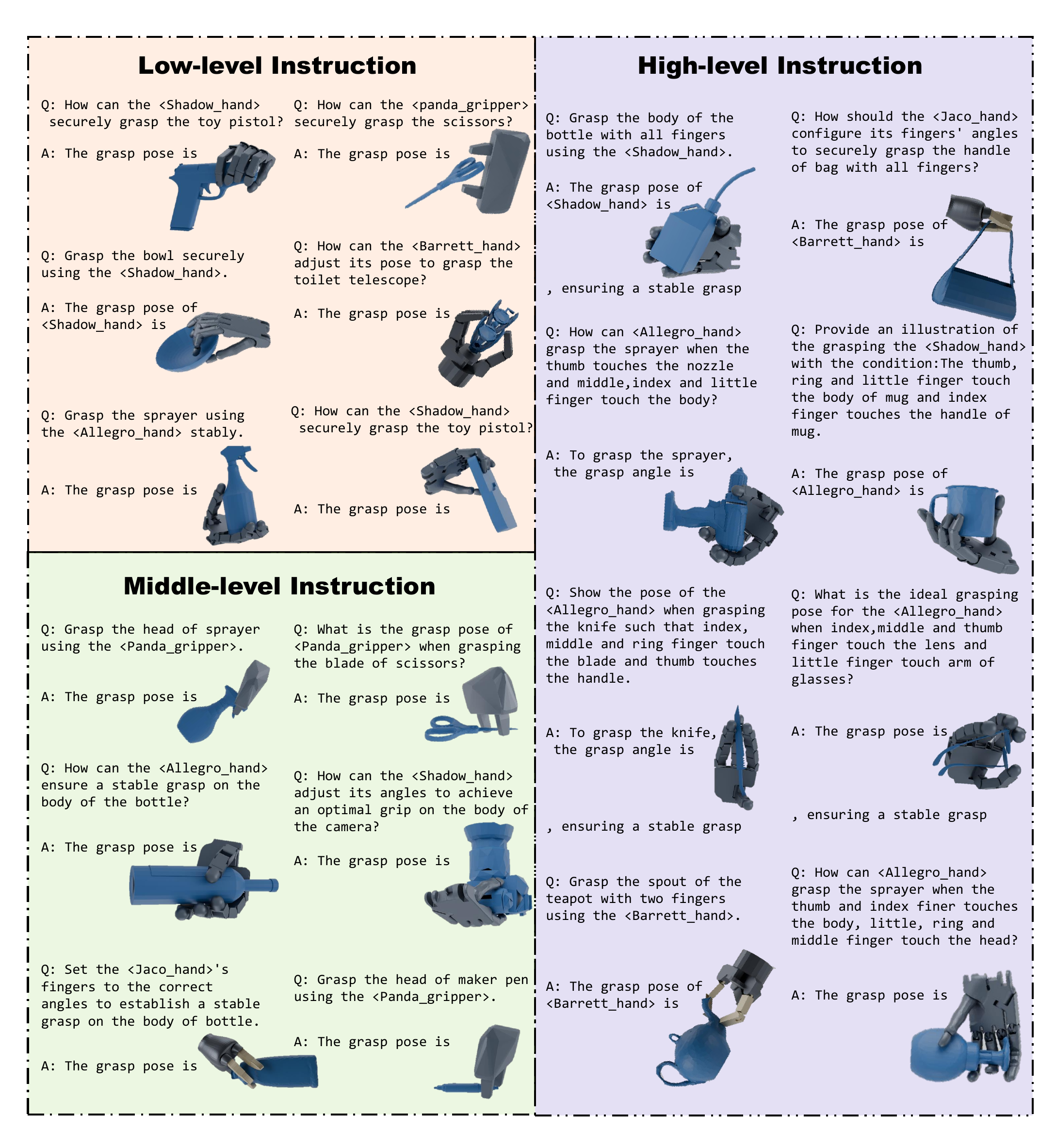}
    
   \caption{Additional visualization of the grasp pose generated by Multi-GraspLLM.}
   \label{vis_graspllm_grasp_supp}
   \vspace{-0.4cm}
\end{figure*}

\begin{table}[h]

\fbox{\parbox{\linewidth}{
    \colorbox{gray!10}{
        \parbox{\dimexpr\linewidth-2\fboxsep\relax}{
        - USER:
Polish the following sentence to make it smooth and fluent, then output the polished sentence. The general meaning of the sentence should not be altered.``\{prompt\}'' Do not change the word ``\{robotics\_hand\}'', and directly output the polished  sentence without quotation marks.
        
        }
    }
}}
    \caption{System prompt of final refinement.}
    \vspace{-0.3cm}
    \label{Refine_Prompt}
\end{table}
For the initial rough template generation, we started with several manually crafted prompts and used GPT-4~\cite{gpt-4} to expand them, as shown in Table~\ref{initial_prompt}. These templates were designed to incorporate essential information, such as robotic hand configurations and grasp details, that vary according to different levels of instruction. Each prompt was structured to query the grasp poses or specify the grasp commands for the target objects as shown in Figure~\ref{vis_graspllm_grasp_supp}. We then populated these templates with contact annotations from Multi-GraspSet.
Due to the complexity of the contact information involved, we further refined the populated templates using GPT-4~\cite{gpt-4} to create the final set of basic conversations, as detailed in Table~\ref{Refine_Prompt}. These iterative refinements helped ensure both clarity and smoothness in the generated dialogue.

    

    

\subsection{Collection of Multi-GraspSet}
We collected 1800 objects from Oakink~\cite{oakink} and 300 objects from ShapeNet~\cite{shapenet} as our dataset base. To facilitate reliable intersection testing, we performed convex decomposition using COACD~\cite{coacd} with a threshold of 0.01 to make the meshes watertight. For grasp pose generation, we used DexGraspNet~\cite{dexgraspnet}, GraspIt~\cite{graspit} and Contact-GraspNet~\cite{contact-graspnet} with their default parameter settings.
After generating grasp poses, we perform penetration checking on all grasp poses, filtering out invalid candidates where penetration distances exceed the threshold (2cm).

For dexterous hands, each object typically corresponds to numerous possible contact patterns due to the complex structure of the dextrous hand. When constructing the conversation dataset, we randomly selected only one grasp pose for each contact pattern.
In contrast, grippers have much simpler mechanics, resulting in only a few possible contact patterns per object. To increase the diversity of gripper grasp pose in our dataset, we randomly selected multiple different grasp poses for each contact information pattern.

\begin{table*}[h]
\setlength{\tabcolsep}{0.5pt}
\begin{center}
\begin{tabular}{l|c c  c|c c c|c  c c|c  c c}
\toprule
\multirow{2}{*}{Method} & \multicolumn{3}{c|}{Low-level} & \multicolumn{3}{c|}{mid-level} & \multicolumn{3}{c|}{high-level} & \multicolumn{3}{c}{Avg} \\

& CD$\downarrow$ & Pen$\downarrow$ & Suc $\uparrow$ & CD $\downarrow$& Pen$\downarrow$ & Suc $\uparrow$& CD$\downarrow$ & Pen$\downarrow$ & Suc$\uparrow$ &CD$\downarrow$ & Pen $\downarrow$& Suc $\uparrow$\\
\midrule
DexGraspNet~\cite{dexgraspnet} &1.05 & 0.79 & 0.37 & 0.99 & 0.76 & 0.29 & 1.02 & 0.79 & 0.34 & 1.02 & 0.78 & 0.33 \\
SceneDiffuser~\cite{Scene-diffusers} & 1.09 & 0.77 & 0.31 & 1.19 & 0.72 & 0.36 & 1.14 & 0.79 & 0.26 & 1.14 & 0.76 & 0.31  \\
DexGYS~\cite{Grasp-as-You-Say} &/&/&/&/&/&/&0.62& 0.68 &0.32&
0.62& 0.68 &0.32\\

Contact-GraspNet~\cite{contact-graspnet} & 0.60 & 0.41 & 0.44  & 0.63 & 0.46 & 0.48 & / &/ & / & 0.62 & 0.43 & 0.46  \\

\rowcolor[gray]{.9} Multi-GraspLLM-mix &0.78 & 0.64 & 0.43 & 0.34 & 0.51 & 0.35 & 0.31 & 0.57 & 0.40  & 0.36& 0.57& 0.40 \\

\toprule

\end{tabular}

\caption{Comparison of baseline method and Multi-GraspLLM in different instruction levels.}
 \vspace{-0.5cm}
\label{levels_result}
\end{center}
\end{table*}

\subsection{Multi-GraspLLM}
We employ PointBERT~\cite{pointbert} as our point cloud encoder, which utilizes the ViT-L/14 model from OpenCLIP~\cite{openclip}. 
For training data, we structure our 1M basic conversations into two parts: 500k conversations for multimodal alignment based on ``Single-Round Grasp Generation'', and another 300k split equally between ``Single-Round Grasp Generation'', ``Multi-Round Mix'', and ``Multi-Round Grasp Generation'' for instruction tuning as mentioned in our main paper. We conduct full-parameter fine-tuning of the LLM backbone and modality adaptor using 8 $\times$ 80G A100 GPUs, while testing is performed on a single A100 GPU.

\vspace{1mm}
\subsection{Evaluation Metrics}
As described in our main paper, we evaluate grasp intention and physical stability using multiple metrics: Chamfer distance, maximum penetration distance, grasp success rate. The Chamfer distance is implemented using PyTorch3D~\cite{pytorch3d}. For our calculations, we uniformly sample 512 points on each robotic hand and utilize the point-to-surface distance through kaolin~\cite{pytorch3d} to calculate the maximum penetration distance.

\section{Additional Ablation Experiments}
\label{Ablation}

We conducted additional experiments on model parameters and instruction levels.
\vspace{1mm}

\vspace{1mm}
\subsection{Effect of Instruction Levels}
As described in the main paper, we establish a three-level instruction hierarchy: low-level instructions without contact information, mid-level instructions specifying only the target grasping part, and high-level instructions containing fine-grained grasp information. Specifically, dextrous hands~\cite{allegro,shadow} consist of all three instruction levels, while the Panda gripper~\cite{panda}, due to its limited degrees of freedom, only operates at mid and low instruction levels. As shown in Table~\ref{levels_result}, Multi-GraspLLM showed distinct differences between different instruction levels, whereas our baseline method without semantic guidance achieves consistent performance.

Our method showed varying performance in terms of Chamfer Distance (CD) which represents grasp intention. Multi-GraspLLM achieved the highest CD in low-level instructions that contain no contact information. For mid-level instructions, our model obtained intermediate CD values. Since high-level instructions contain the most detailed contact information, our model demonstrated the lowest CD, which also indicates the best grasp intention consistency. Notably, other grasp quality metrics remained consistent across all levels. These results align with our intuitive expectations.

\begin{figure}[th]
    \includegraphics[width=\linewidth]{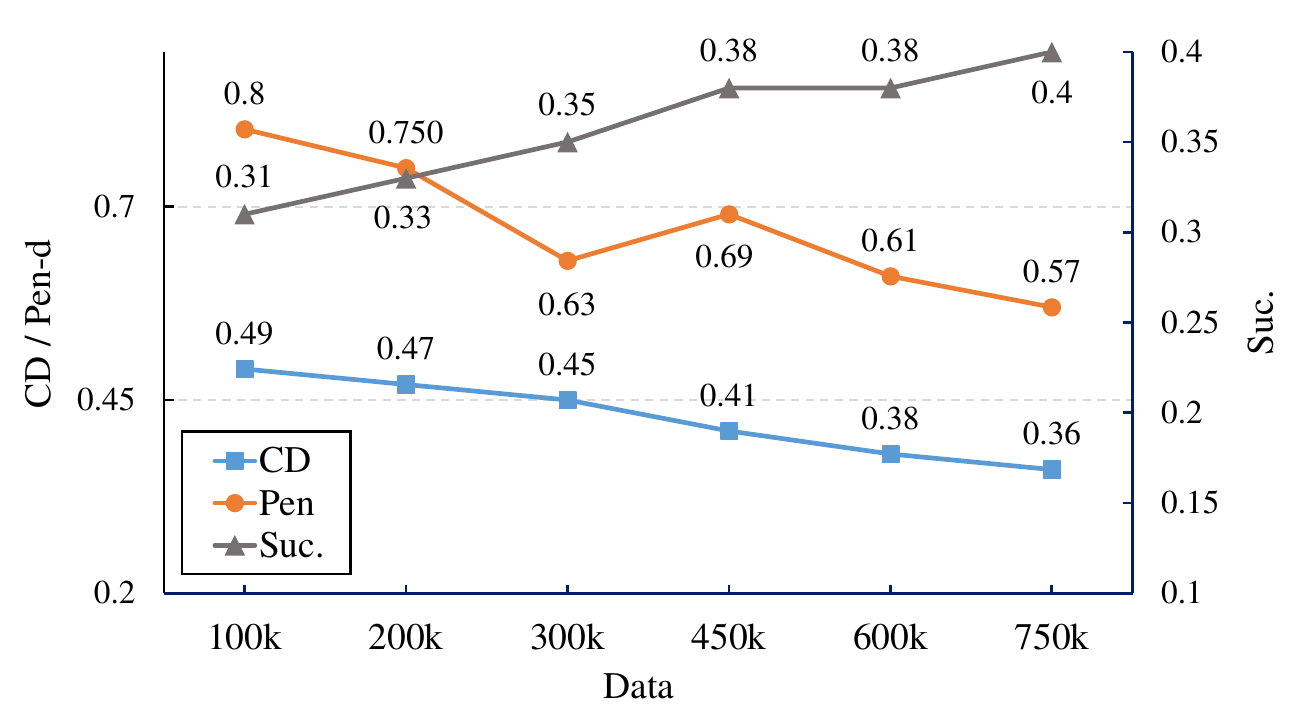}
   \caption{Ablation on training data for modality alignment.}
   \label{plot}
   \vspace{-0.2cm}
\end{figure}

\vspace{1mm}
\subsection{Effect of Training Data Amount for Modality Alignment} 
We also analyzed the effect of the size of the training data in modality alignment and structure of data in instruction tuning, as shown in Figure~\ref{plot} and Table~\ref{table_ablation3}. 
We gradually reduced the amount of modality alignment data to test our alignment method. As shown in Figure \ref{plot}, using more data improves the performance of our model in intention understanding (lower CD) and grasp quality (higher Suc).

\begin{table}[h]
\setlength{\tabcolsep}{3pt}

\begin{center}
\begin{tabular}{l|c|c |c| c }
\toprule
 Number& CD$\downarrow$ & Pen$\downarrow$ & Suc $\uparrow$  \\
\midrule
One hand	&0.41	&1.14 &	0.32\\
Two hands	&0.43	&1.23	&0.29\\
Three hands	&0.41	&1.11&	0.31\\
Four hands	&0.39	&1.02&	0.33\\
Five hands	& \textbf{0.37}	&\textbf{1.00}	&\textbf{0.34}\\

\midrule
\end{tabular}
\caption{Performance of Allegro hand in different number of hands used in training. }
\label{table_ablation4}
\vspace{-0.8cm}
\end{center}
\end{table}
\vspace{1mm}

\begin{figure*}[h]
  \centering
    \includegraphics[width=\linewidth]{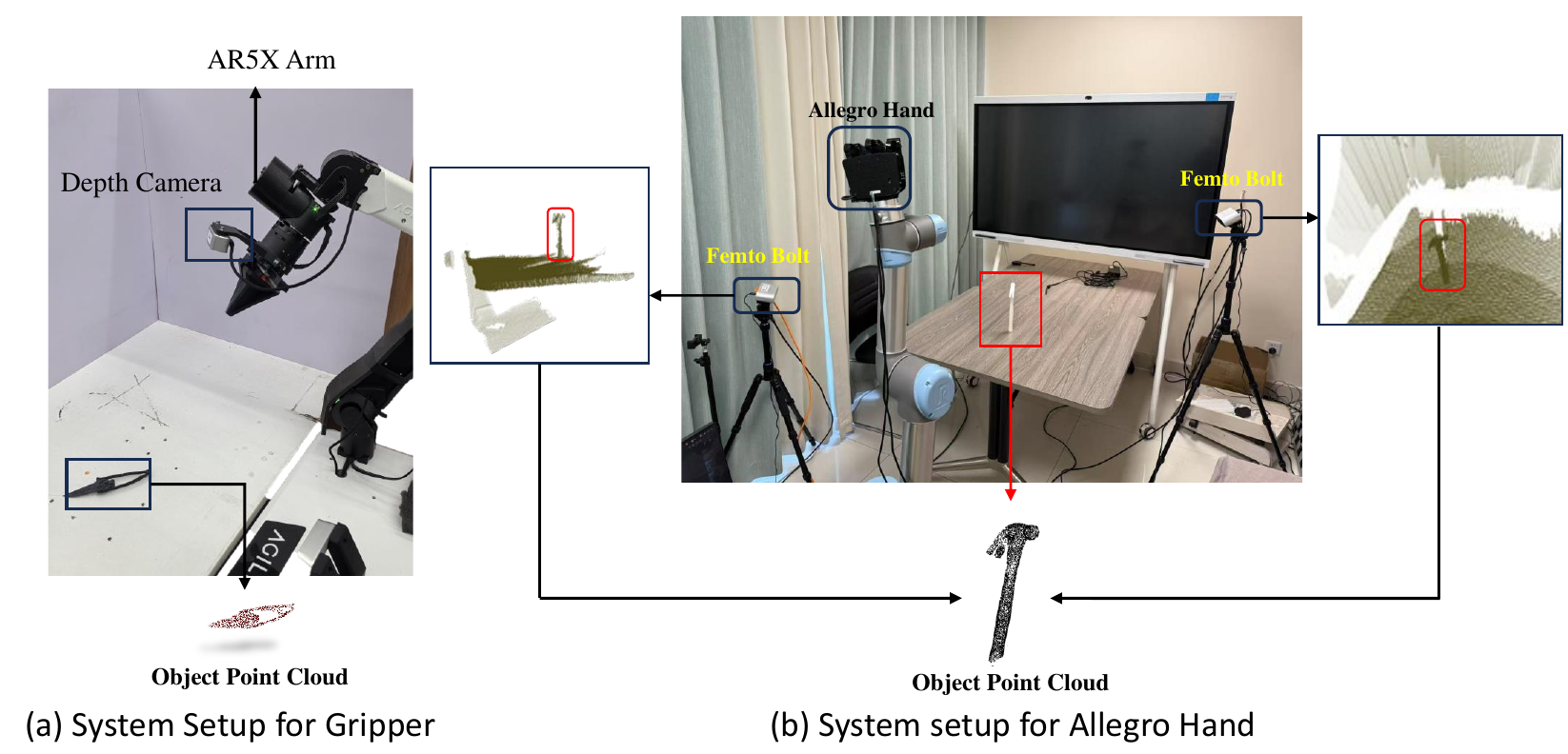}
    
   \caption{Robot experiment setup for Allegro.}
   \label{Robo_sys}
\end{figure*}
\subsection{Effect of Used Hands in Training} 
Finally, we analyzed the impact of the used hands involving in training. We evaluated the performance of the Allegro hand as the total number of training hands changed. As shown in Table~\ref{table_ablation4}, ``Two hands'' stands for Allegro and Shadow, ``Three hands'' means Allegro, Shadow, Panda,  ``Four hands'' means Allegro, Shadow, Panda, and Barrett  and ``Five hands'' means Allegro, Shadow, Panda, Barrett, and Jaco. Starting from three hands, the mixed training results were already better than the separate training results. Moreover, as the number of hands increases and the data volume grows, the model's performance improves.

    

\section{Real World Experiments}
\label{Real}
We conducted extensive experiments in real world environments. 

\vspace{1mm}
\subsection{Robot System}
As shown in the Figure~\ref{Robo_sys}, 
We built two systems in a real environment to test the model's performance on a gripper and a multi-fingered dexterous hand. As shown in the left part of Figure~\ref{Robo_sys}, our robot gripper setup consists of an AR5X arm with a gripper and an Orbbec DaBai depth camera. We operate the robotic arm at different positions and capture various depth images using the camera at the wrist.
Our dexterous hand system is composed of a UR5 arm and an Allegro hand, with two Orbbec Femto Bolt depth cameras, as shown in the right part of Figure~\ref{Robo_sys}. In both systems, we obtain full object point clouds from the depth images with the help of SAM2~\cite{sam2}.


We collect 29 objects from daily life covering various objects from tools to potted plants.
We tested both the baseline method for grippers~\cite{contact-graspnet,6-dof-graspnet} and Multi-GraspLLM on these objects, evaluating them based on grasping success rate (Suc) and part selection accuracy (Acc). For each method, we conducted 20 tests and calculated the average performance, with both part selection correctness and grasping success being assessed manually. For both systems, we filtered out all grasp poses that have collisions with the table for evaluation.

    

\vspace{1mm}

\begin{table}[h!]
\setlength{\tabcolsep}{4pt}

\begin{center}

\begin{tabular}{l|c c |c c}
\toprule
\multirow{2}{*}{method} & \multicolumn{2}{c|}{Gripper} & \multicolumn{2}{c}{Allegro}  \\
&Suc & Acc & Suc & Acc\\

\midrule

Contact-GraspNet~\cite{contact-graspnet}& 0.48  & 0.42 &/&/  \\

Scenediffers~\cite{Scene-diffusers}&/&/ &0.24&0.30  \\
DexGraspNet~\cite{dexgraspnet}&/&/ &0.29&0.32  \\
DexGYS~\cite{Grasp-as-You-Say}&/&/ &0.27&0.56  \\

Multi-GraspLLM& \textbf{0.45} & \textbf{0.82} & \textbf{0.31} & \textbf{0.72} \\
\midrule
\end{tabular}
\caption{Result of real world experiments. }
\label{table_real}
\vspace{-0.4cm}
\end{center}
\end{table}

\subsection{Experiments Results}
\begin{figure*}[h]
   \centering
    \includegraphics[width=0.85\linewidth]{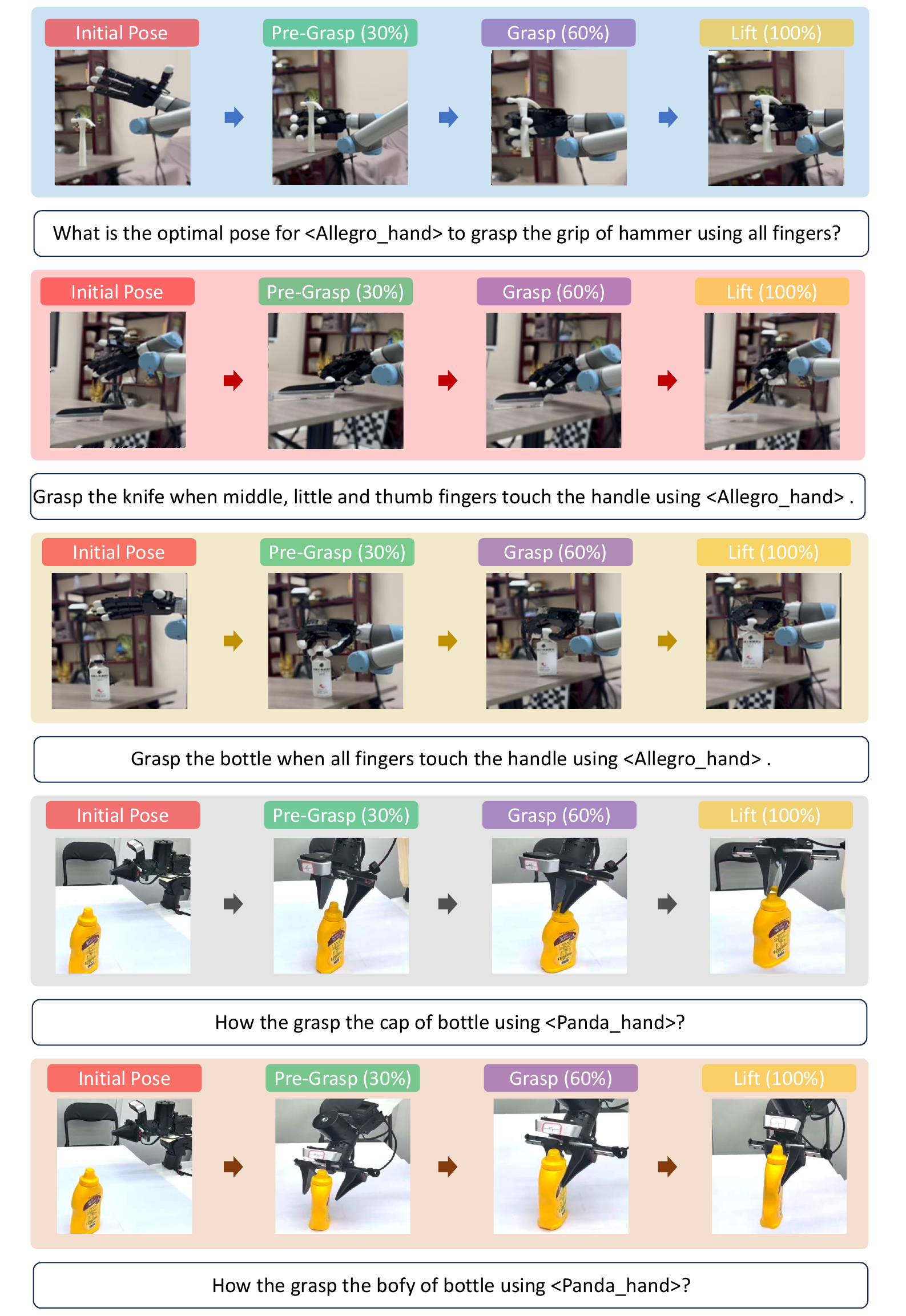}
    
   \caption{Illustration of grasp process in real world.}
   \label{Real_grasp}
   \vspace{-0.4cm}
\end{figure*}

As shown in Table~\ref{table_real}, our Multi-GraspLLM achieves comparable grasp success rates to traditional methods that focus on grasp quality and DexGYS~\cite{Grasp-as-You-Say} that use semantic guidance, while significantly outperforming them in part selection accuracy, which measures grasp intention. These results demonstrate both the sim-to-real feasibility of our approach and its superiority in grasp intention understanding. We show the visualization of grasping experiments under different grasp descriptions in Figure~\ref{Real_grasp}.


\end{document}